%% file: main.tex
\documentclass{article}
\usepackage{spconf,amsmath,graphicx}

\title{Rethinking temporal self-similarity for repetitive action counting}
\name{Yanan Luo*$^{1}$, Jinhui Yi*$^{1}$, Yazan Abu Farha$^{2}$, Moritz Wolter$^{1}$, Juergen Gall$^{1,3}$ 
\thanks{
This work has been supported by the project iBehave (receiving funding from the programme “Netzwerke 2021”, an initiative of the Ministry of Culture and Science of the State of Northrhine Westphalia), the ERC Consolidator Grant FORHUE (101044724), and the Chinese Scholarship Council (202108440041). * denotes equal contribution.}
}
\address{$^{1}$University of Bonn \quad $^{2}$Birzeit University \quad $^{3}$Lamarr Institute for Machine Learning and Artificial Intelligence}
\usepackage{hyperref}

\usepackage{amsmath}
\usepackage{algorithm}
\usepackage{algpseudocode}

\usepackage{graphicx}
\graphicspath{ {./figures/} }

\usepackage{standalone}
\usepackage{tikz}
\usepackage{pgfplots}

\definecolor{darkgray176}{RGB}{176,176,176}
\definecolor{darkorange25512714}{RGB}{255,127,14}
\definecolor{forestgreen4416044}{RGB}{44,160,44}
\definecolor{steelblue31119180}{RGB}{31,119,180}

\usepackage{booktabs}
\usepackage{multirow}

\usepackage{enumitem}

\usepackage{caption}
\usepackage{subcaption}

\usetikzlibrary{shapes.geometric}
\usepackage{siunitx}
\usepackage{tablefootnote}

\usepackage{comment}

\begin{document}
%\ninept
%
\maketitle
\input{sec/0_abstract}
\input{sec/1_intro}
\input{sec/2_related}
\input{sec/3_method}

\input{sec/4_experiments}

\input{sec/5_conclusion}

\end{document}

%% file: sec/0_abstract.tex
\begin{abstract}
% Problem
% Counting repetitive actions in long untrimmed videos is a challenging task that has many applications such as rehabilitation. 
% State-of-the-art methods predict action counts by first generating a temporal self-similarity matrix (TSM) from the sampled frames and then feeding the matrix to a predictor network. The self-similarity matrix, however, is not an optimal input to a network   
% While such methods achieve good accuracy, their performance is sub-optimal as they only consider the TSM for predicting the action counts. 
% Furthermore, these approaches usually work with very few frames as input, which leads to the destruction of the complete action cycle. 
% In this paper, we propose a framework for repetition counting that takes all the video frames as input with full temporal resolution. 
% The proposed framework predicts action start probabilities and derives the action count by counting the number of frames that correspond to an action repetition start. In contrast to current approaches that have the TSM as a bottleneck, we generate the reference TSM as an auxiliary task and force it to capture the repetitive nature of the action for temporal repetition consistency.
% The proposed framework achieves state-of-the-art or comparable results with the metric of MAE, OBO on three datasets, i.e., RepCount, UCFRep, and Countix.
Counting repetitive actions in long untrimmed videos is a challenging task that has many applications such as rehabilitation. 
State-of-the-art methods predict action counts by first generating a temporal self-similarity matrix (TSM) from the sampled frames and then feeding the matrix to a predictor network. The self-similarity matrix, however, is not an optimal input to a network since it discards too much information from the frame-wise embeddings. We thus rethink how a TSM can be utilized for counting repetitive actions and propose a framework that learns embeddings and predicts action start probabilities at full temporal resolution. The number of repeated actions is then inferred from the action start probabilities. In contrast to current approaches that have the TSM as an intermediate representation, we propose a novel loss based on a generated reference TSM, which enforces that the self-similarity of the learned frame-wise embeddings is consistent with the self-similarity of repeated actions. The proposed framework achieves state-of-the-art results on three datasets, i.e., RepCount, UCFRep, and Countix.

% While such methods achieve good accuracy, their performance is sub-optimal as they only consider the TSM for predicting the action counts. 
% Furthermore, these approaches usually work with very few frames as input, which leads to the destruction of the complete action cycle. 
% In this paper, we propose a framework for repetition counting that takes all the video frames as input with full temporal resolution. 
% The proposed framework predicts action start probabilities and derives the action count by counting the number of frames that correspond to an action repetition start. In contrast to current approaches that have the TSM as a bottleneck, we generate the reference TSM as an auxiliary task and force it to capture the repetitive nature of the action for temporal repetition consistency.
% The proposed framework achieves state-of-the-art or comparable results with the metric of MAE, OBO on three datasets, i.e., RepCount, UCFRep, and Countix.
\end{abstract}

%% file: sec/1_intro.tex
\section{Introduction}
\label{sec:intro}

\begin{figure}[t]
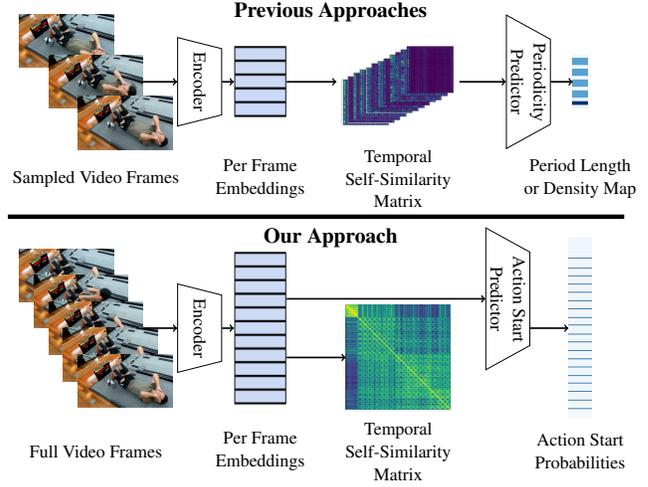
%[htbp!]
    \centering
    
    \includestandalone[width=0.98\columnwidth]{figures/intro_img_1}
    \vspace{-2mm}
    \caption[caption]{
    While previous works use TSMs as intermediate representation, we learn a representation where the TSM is consistent with the ground-truth.  
    %Comparison of state-of-the-art approaches and our proposed approach. 
    %State-of-the-art repetitive action counting approaches use the Temporal Self-similarity Matrix (TSM) as the bottleneck to predict action count. This can result in a loss of fine-grained information. In contrast to these approaches, our prediction network works directly on the frame-wise embeddings to predict frame-wise action repetition start probabilities, from which we count the number of repetitions. Furthermore, our framework generates the TSM as an auxiliary task and does not have a bottleneck. 
    }
    \label{fig:teaser}
    \vspace{-5mm}
\end{figure}

% bg & 3 characteristics
Repetitive actions are ubiquitous in the real world, ranging from natural phenomena like glacial and oceanic currents to fundamental biological processes like human heartbeat~\cite{gideon2021way, fieraru2021aifit, zhang2021repetitive}. In this paper, we address the task of class-agnostic repetitive action counting, wherein the goal is to predict the number of repetitions of an action that are carried out in a video. Such a task has many practical applications such as guiding people during their physical exercises. 

%In many real-world scenarios, repetitive actions are not performed at a fixed frequency. For example, the duration of each squat depends on the athlete's strength and stamina, which cannot be measured by a single regular cycle. 

In order to count repetitive actions, \cite{dwibedi2020counting} introduced the temporal self-similarity matrix (TSM) which computes similarities between frame-wise features.
As shown in the top row of Fig.~\ref{fig:teaser}, the TSM is used as intermediate representation to predict frame-wise period length and classify per-frame binary periodicity jointly, which are then merged for counting. 
%As shown in the top row of Figure~\ref{fig:teaser}, the TSM is used as input to a network to predict frame-wise period length and classify per-frame binary periodicity jointly, which are then merged for counting. 
%\cite{hu2022transrac,li2023full} followed this paradigm and proposed stronger decoders to obtain better estimates. 
\cite{hu2022transrac,li2023full} extended this paradigm and learn multiple correlation matrices. 
For all these approaches, the frame-features are reduced to TSMs and the periodicity or a density map are estimated from the TSMs. 
While self-similarity of features is an important cue for repeating actions, the TSM discards too much information from the frame-wise features and it is thus not the optimal input for a prediction network. We thus rethink the way how a temporal self-similarity matrix is used for class-agnostic repetition counting and propose a different approach of utilizing self-similarity as shown in the bottom row of Fig.~\ref{fig:teaser}.    

Instead of reducing features to a TSM and predicting a density map of actions from the TSM, we predict the start of each repeated action from the frame-wise embeddings directly. In order to learn embeddings where the self-similarity is high for repeated actions, we introduce a new loss that enforces consistency between the self-similarity of the learned frame-wise embeddings and the self-similarity of repeated actions. To this end, we construct a target TSM from the ground-truth annotations and aim at a high embedding similarity when actions are repeated. We evaluate the approach on three datasets, namely RepCountA \cite{hu2022transrac}, UCFRep \cite{Zhang_2020_CVPR}, and Countix \cite{dwibedi2020counting}, where our approach achieves state-of-the-art results. In particular, in terms of Off-By-One Accuracy (OBOA), our approach outperforms the state of the art by a large margin.     
Overall, the contributions of this work can be summarized as follows:
%\vspace{-3pt}
\begin{itemize}[itemsep=-1pt]%[nosep]%[topsep=0pt]
    \item We propose a framework for repetitive action counting that works on the full temporal resolution and does not have a TSM bottleneck.
    %\item We define the generation of the reference TSM as an auxiliary task and force it to capture the repetitive action structure.
    \item We introduce a novel loss based on a generated reference TSM to enforce the self-similarity of the learned frame-wise embeddings to be consistent with the self-similarity of repeated actions. 
    %temporal repetition consistency loss, enhancing the feature learning process to acquire repetition-sensitive frame-wise features.    
    %\item Our proposed approach achieves state-of-the-art results on three datasets: RepCount, UCFRep, and Countix.
\end{itemize}

%% file: figures/intro_img_1.tex
% This file was created with tikzplotlib v0.10.1.
\begin{tikzpicture}
\tikzstyle{every node}=[font=\LARGE]
\tikzset{
        box/.style={
        rectangle,
        minimum width=0.5cm,
        minimum height=0.5cm
        }
    }

    % Figure 1
    \node[inner sep=0pt] (i1) at (-1,-1)
    {\includegraphics[width=.18\textwidth]{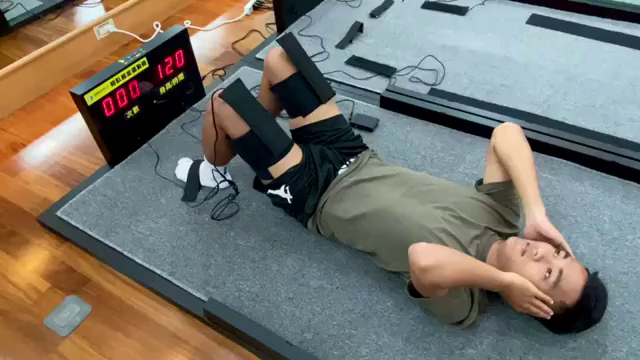}};
    \node[inner sep=0pt] (i2) at (0,-2.4) % -3.8
    {\includegraphics[width=.18\textwidth]{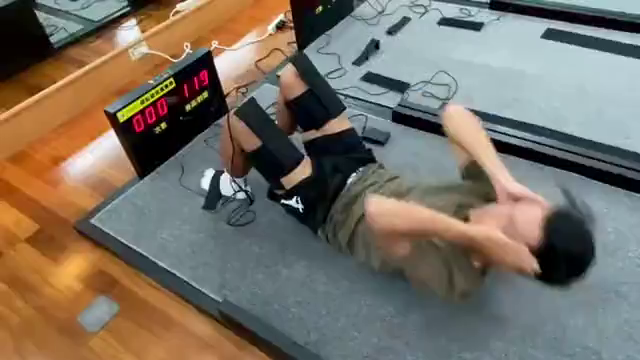}};
    \node[inner sep=0pt] (i3) at (1,-3.8) % -5.7 -7.6
    {\includegraphics[width=.18\textwidth]{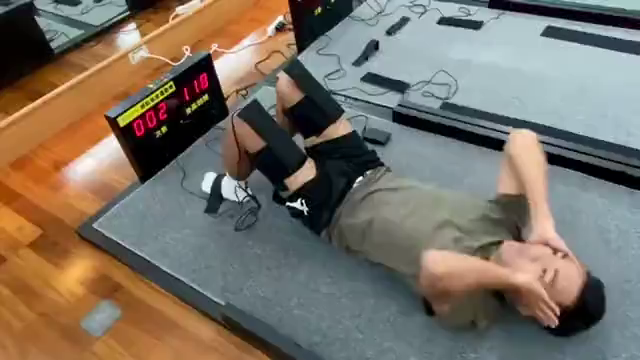}};
    \node[box==text, align=center](a) at (0, -5.7){ Sampled Video Frames}; % -9

    \node[trapezium,
    draw,
    thick,
    trapezium stretches = true,
    minimum width = 3cm,
	minimum height = 1.5cm,
    trapezium angle=60,
    rotate=270,
    anchor=north 
    ] (t1) at (4.3, -2.4) {Encoder};
    \draw [->, ultra thick] (i2.east|-t1.south) -- (t1.south);

    \node[inner sep=0pt] (pfembed) at (5.6,-2.4)
    {\includegraphics[width=.1\textwidth]{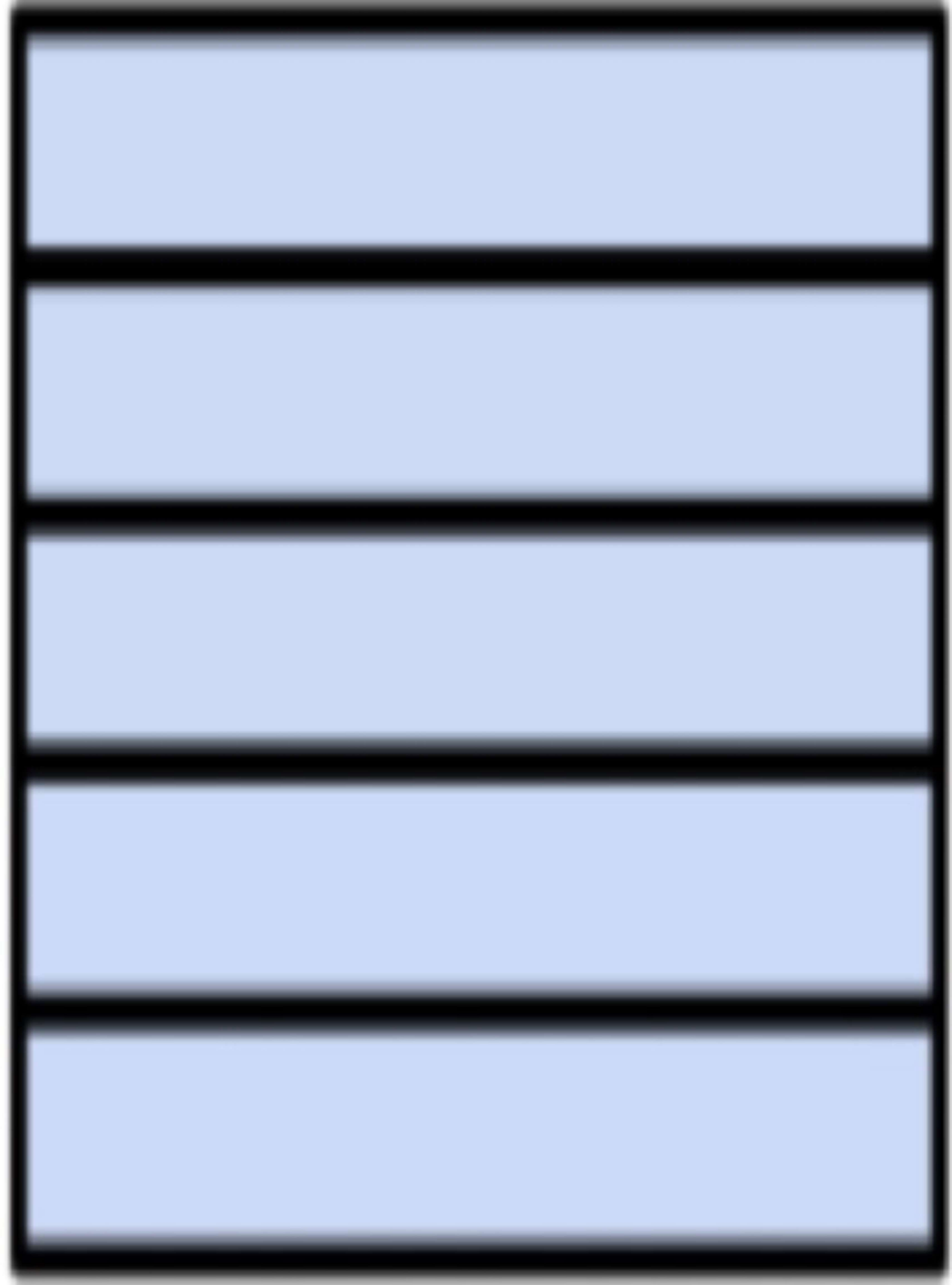}};
    \node[box==text, align=center](b) at (5.6, -5.7){Per Frame\\Embeddings};
    \draw [->, ultra thick] (t1.north|-pfembed.west) -- (pfembed.west);

    \node[inner sep=0pt] (tsm1) at (9.2,-3.1)
    {\includegraphics[width=.11\textwidth]{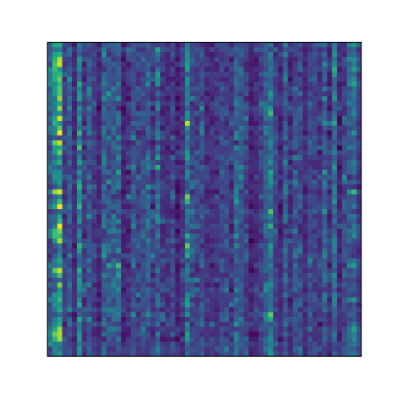}};
    \node[inner sep=0pt] (tsm2) at (9.4,-3.0)
    {\includegraphics[width=.11\textwidth]{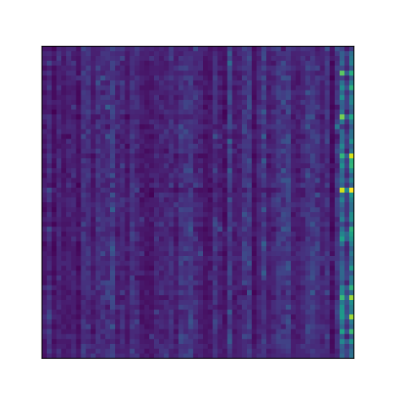}};
    \node[inner sep=0pt] (tsm3) at (9.6,-2.9)
    {\includegraphics[width=.11\textwidth]{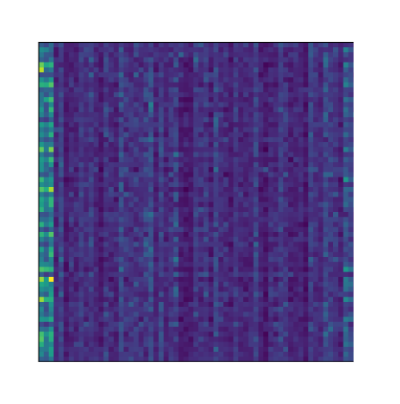}};
    \node[inner sep=0pt] (tsm4) at (9.8,-2.8)
    {\includegraphics[width=.11\textwidth]{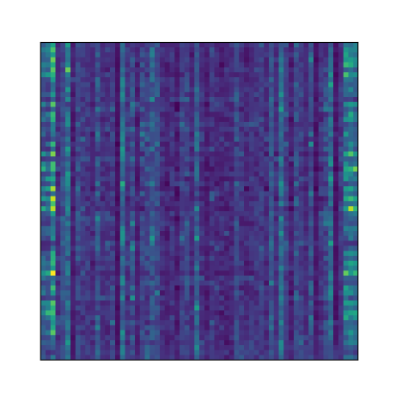}};
    \node[inner sep=0pt] (tsm5) at (10.0,-2.7)
    {\includegraphics[width=.11\textwidth]{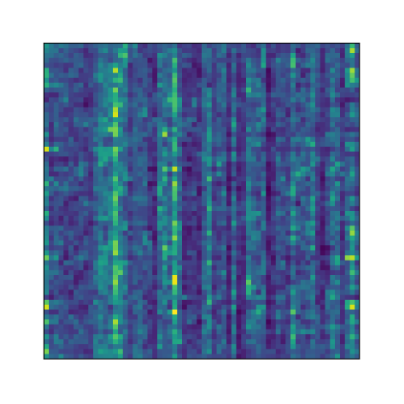}};
    \node[inner sep=0pt] (tsm6) at (10.2,-2.6)
    {\includegraphics[width=.11\textwidth]{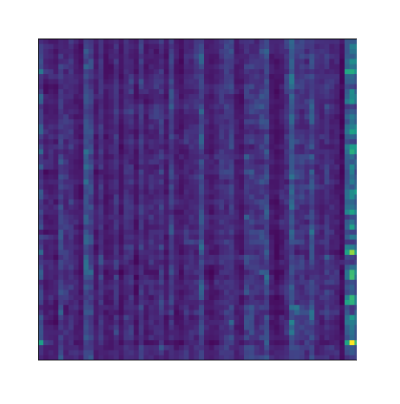}};
    \node[inner sep=0pt] (tsm7) at (10.4,-2.5)
    {\includegraphics[width=.11\textwidth]{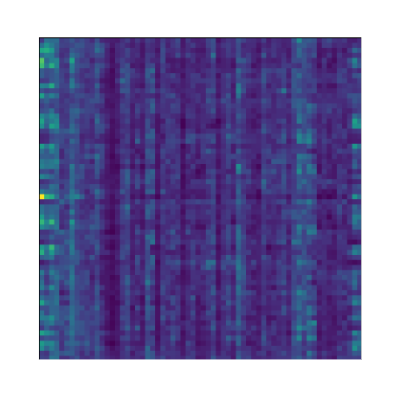}};
    \node[inner sep=0pt] (tsm8) at (10.6,-2.4)
    {\includegraphics[width=.11\textwidth]{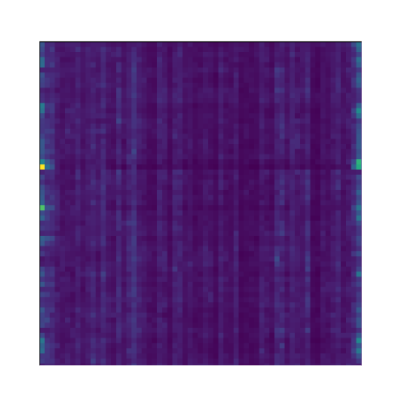}};
    \node[inner sep=0pt] (tsm9) at (10.8,-2.3)
    {\includegraphics[width=.11\textwidth]{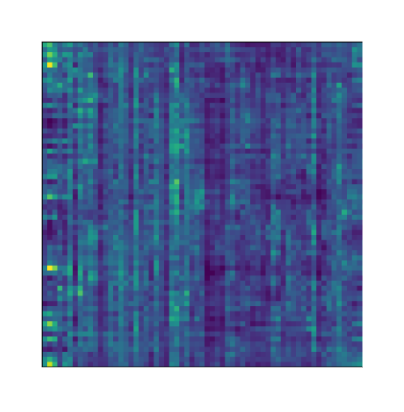}};
    \node[inner sep=0pt] (tsm10) at (11.0,-2.2)
    {\includegraphics[width=.11\textwidth]{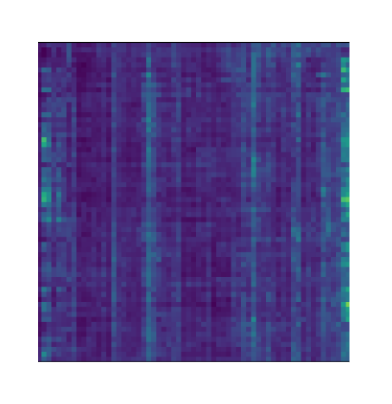}};
    \node[inner sep=0pt] (tsm11) at (11.2,-2.1)
    {\includegraphics[width=.11\textwidth]{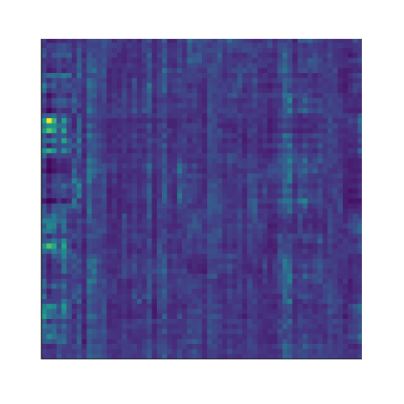}};
    \node[inner sep=0pt] (tsm12) at (11.4,-2.0)
    {\includegraphics[width=.11\textwidth]{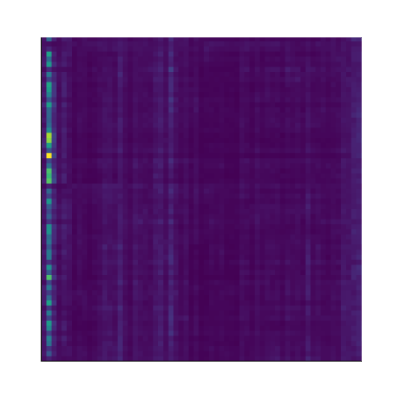}};
    
    \node[box==text, align=center](c) at (10.3, -5.7){Temporal\\Self-Similarity\\Matrix};
    \draw [->, ultra thick] (6.5, -2.4) -- (8.2, -2.4);
    %(pfembed.east|-tsm1.west) -- (6.6, -2.4) -- (8.6, -2.4) -- (tsm1.west);

    \node[trapezium,
    draw,
    thick,
    trapezium stretches = true,
    minimum width = 0.1cm,
    minimum height = 1.5cm,
    trapezium angle=60,
    rotate=270,
    anchor=north,
    align=center
    ] (t2) at (15.5, -2.4) {Periodicity \\Predictor};
    \draw [->, ultra thick] (tsm12.east|-t2.south) -- (t2.south);

    \node[inner sep=0pt] (final) at (16.5, -2.4) {
        \includegraphics[scale=0.25]{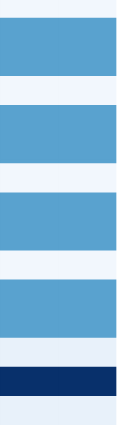}
    };
    \node[box==text, align=center](e1) at (16.5, -5.7){Period Length\\or Density Map};

    \draw [->, ultra thick] (t2.north|-final.west) -- (final.west);

    \node[box==text](h1) at (8, 0){\huge\textbf{Previous Approaches}};

    % Mid line
    \draw [line width=1.5mm,   black] (-3,-7) -- (18,-7) node [right] {};

    % Figure 2

    \node[box==text](h2) at (8, -7.7){\huge \textbf{Our Approach}};

    \node[inner sep=0pt] (i12) at (-1,-9.0)
    {\includegraphics[width=.18\textwidth]{./images_intro/plot1/0.png}};
    \node[inner sep=0pt] (i22) at (-0.5,-9.9)
    {\includegraphics[width=.18\textwidth]{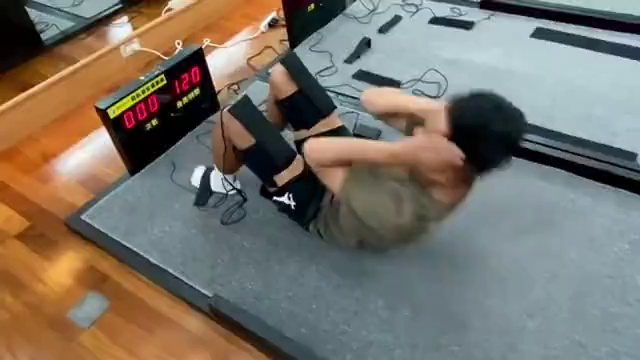}};
    \node[inner sep=0pt] (i32) at (0,-10.8)
    {\includegraphics[width=.18\textwidth]{./images_intro/plot1/123.png}};
    \node[inner sep=0pt] (i42) at (0.5,-11.7)
    {\includegraphics[width=.18\textwidth]{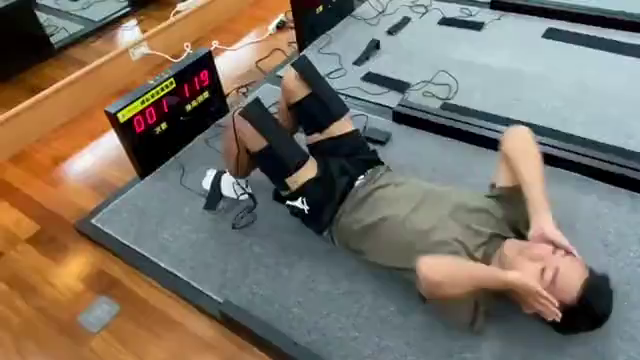}};
    \node[inner sep=0pt] (i52) at (1,-12.6)
    {\includegraphics[width=.18\textwidth]{./images_intro/plot1/166.png}};
    \node[box==text](a2) at (0, -15){\LARGE Full Video Frames};

    \node[trapezium,
    draw,
    thick,
    trapezium stretches = true,
    minimum width = 3cm,
	minimum height = 1.5cm,
    trapezium angle=60,
    rotate=270,
    anchor=north
    ] (t3) at (4.3, -10.8) {Encoder};
    \draw [->, ultra thick] (i32.east|-t3.south) -- (t3.south);

    \node[inner sep=0pt] (pfembed1) at (5.6,-10.8)
    {\includegraphics[width=.1\textwidth]{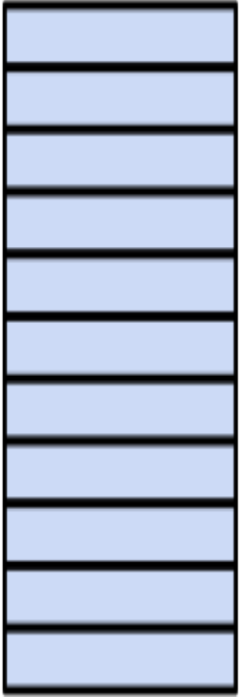}};
    \node[box==text, align=center](b2) at (5.6, -15){Per Frame\\Embeddings};
    \draw [->, ultra thick] (t3.north|-pfembed1.west) -- (pfembed1.west);

    \node[trapezium,
    draw,
    thick,
    trapezium stretches = true,
    minimum width = 3cm,
	minimum height = 1.5cm,
    trapezium angle=60,
    rotate=270,
    anchor=north,
    align=center
    ] (t3) at (14.8, -9.8) {Action Start \\Predictor};
    \draw [->, ultra thick] (pfembed1.east|-t3.south) -- (t3.south);

    \node[inner sep=0pt] (tsm22) at (10.3,-11.8)
    {\includegraphics[width=.2\textwidth]{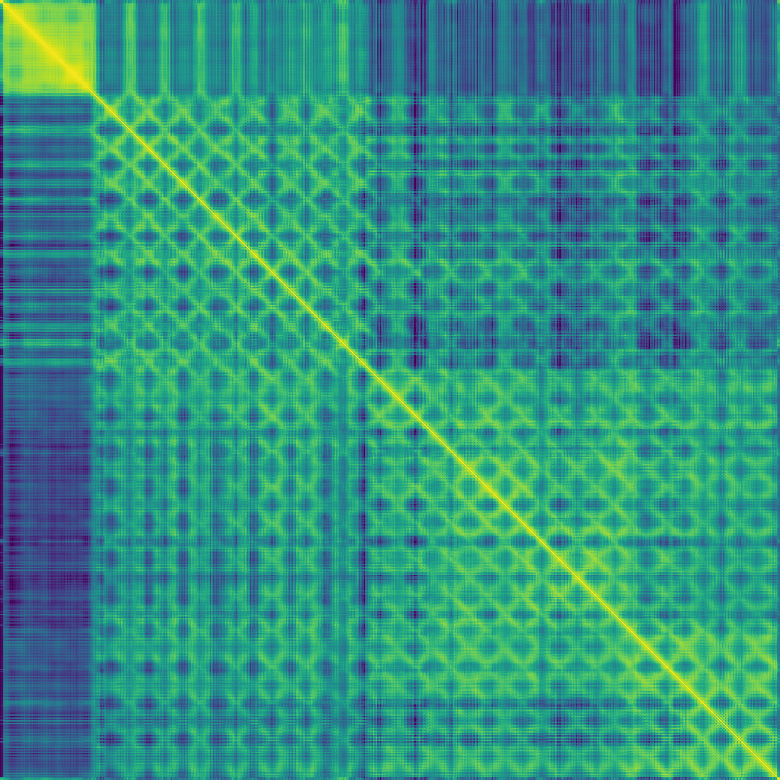}};
    \node[box==text, align=center](c2) at (10.3, -15){Temporal\\Self-Similarity\\Matrix};
    \draw [->, ultra thick] (pfembed.east|-tsm22.west) -- (tsm22.west);
    
    \node[inner sep=0pt] (final) at (16.5, -10.8) {
        \includegraphics[scale=0.65]{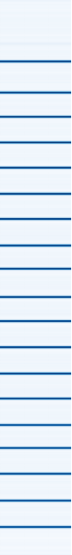}
    };
    \node[box==text, align=center](e2) at (16.5, -15){Action Start\\Probabilities};
    \draw [->, ultra thick] (t3.north|-final.west) -- (final.west);
\end{tikzpicture}

%% file: sec/2_related.tex
\section{Related work}

%\textbf{Equidistant and variable Periods.}
Early action counting models were built with hard-coded constraints and assumed a uniform action period~\cite{vlachos2005periodicity}. Their modeling process typically involved a two-stage approach. The first step encompassed the segmentation and tracking of objects~\cite{satkin2010modeling, pogalin2008visual, vlachos2005periodicity, liu1998finding} within the video frames, effectively isolating the objects of interest. Subsequently, spectral or frequency feature engineering techniques~\cite{briassouli2007extraction, cutler2000robust} were applied to the segmented objects \cite{panagiotakis2018unsupervised}, generating various outputs~\cite{pogalin2008visual, thangali2005periodic, cutler2000robust} for counting. These early methods are limited to action cycles of equal length, which rarely happens in real-world scenarios. 
% various periods

To detect various periodicity, previous works~\cite{levy2015live, dwibedi2020counting, hu2022transrac} proposed to take input video frames at different time scales. Despite the efficacy demonstrated by these approaches, they are constrained by the limited number of frames as input, rendering them impractical for processing extended video sequences. 
The recent approach \cite{li2023full} uses full-resolution as input, but it relies on a strided convolution of the input video. Technically, their network does not see a full-resolution video for training. Multi-striding or scaling makes inference complicated, and the temporal information from the original data is lost due to sampling. 
In contrast to these approaches, our work focuses on full resolution without any sort of frame sampling to estimate the repetitive actions more accurately. The concurrent work~\cite{qiu2024multipath} integrates object detection and multi-path transformers to regress the density map.
\cite{dwibedi2020counting} proposed the RepNet architecture which relies on a Temporal Self-similarity Matrix (TSM)~\cite{junejo2010view, benabdelkader2001eigengait} as its only intermediate layer. Following this line of research, \cite{hu2022transrac} proposed the TransRAC architecture which constructs multi-scale-sample inputs and regresses the density map from the intermediate correlation matrices constructed by adopting multi-head attention~\cite{vaswani2017attention}. Based on TransRAC, \cite{li2023full} proposed to compute correlation matrices from refined features extracted by a temporal convolution network (TCN). 
%encode the 5-step-sample frames, which followed the correlation matrix by a TCN (temporal convolution network), the authors obtained state-of-the-art results. 
%However, it relies on a strided convolution of the input video. Technically their network does not see a full-resolution video for training. 
%Multi-striding or scaling makes inference more complicated. Additionally, temporal information from the original data is lost. 
However, using a TSM or correlation matrices as an exclusive intermediate layer causes information loss and introduces a bottleneck. Different from previous approaches, we learn embeddings where the temporal self-similarity is consistent with the structure of the repetitions in the input video.

%Unlike previous methods that perform action counting from a bottleneck of the model, \textit{i.e.} TSM, we focus on temporal repetition consistency learning in videos. 
%So we proposed a framework named RacNet that takes full-resolution video frames as input 
%and predicts action start probability 
%by adopting a multi-stage temporal convolutional network(MS-TCN)\cite{farha2019ms}. 
%Meanwhile, construct a repetition consist ground truth TSM to utilize the prior repetition action structure in videos. Train with the temporal repetition constraint loss between predicting TSM and generating TSM to enhance the repetition consistency learning into feature learning. 
%We use full resolution without stride or any other pre-processing to train the model and predict the counts. Thus for inference, we can feed any data without considering the numbers of the sequences, which is more flexible, simple, but effective. 

%% file: sec/3_method.tex
\section{RACnet model}
\label{sec:racnet}

\begin{figure*}[t]
    \centering
    \includegraphics[width = 0.85\textwidth]{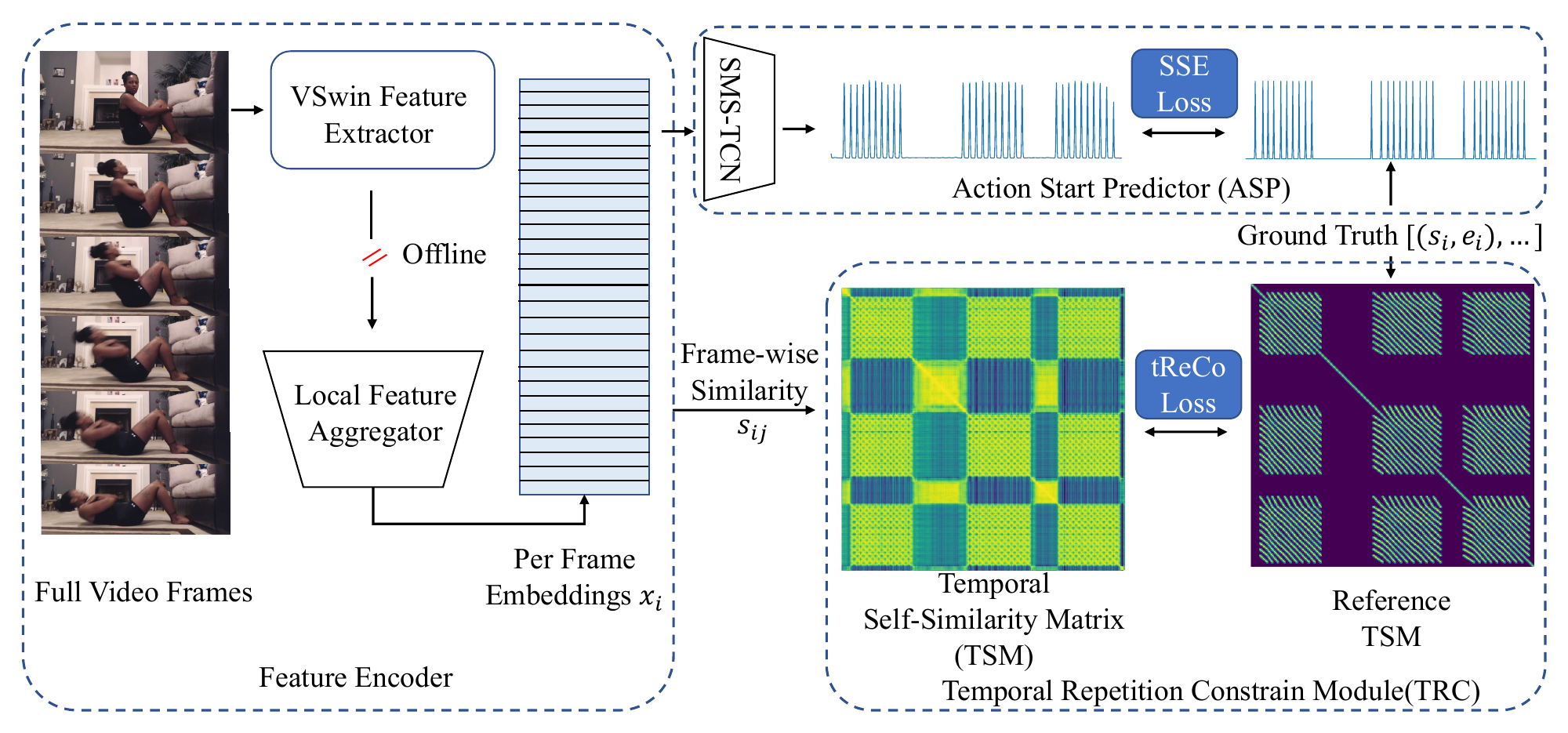} 
    \vspace{-4mm}
    \caption[caption]{\textbf{RACnet architecture}. Our approach consists of three modules: The \textbf{Feature Encoder} generates per-frame embeddings from full-resolution videos. It is pre-trained and frozen. 
    % where the pipeline before the red symbol is offline.
    %where the feature extraction is performed offline. 
    The \textbf{Temporal Repetition Constrain Module} generates a temporal self-similarity matrix as an auxiliary task, where the \textbf{t}emporal \textbf{Re}petition \textbf{Co}nstrain (tReCo) loss is proposed to enforce consistency between the self-similarities of the features and the repeated actions. The \textbf{SMS-TCN} network in the \textbf{Action Start Predictor (ASP)} generates per-frame action start probabilities and the sum-of-squared loss (SSE) is used for training. The number of repetitions in the input videos is calculated as the number of frames that correspond to an action start.}
    \label{fig:racnet}
    \vspace{-5mm}
\end{figure*}

In this section, we introduce our Repetitive Action Counting network (RACnet). We cast the action counting task as action start prediction. During inference, the action count can be directly estimated by counting the number of action starts that are predicted by the network. 
Given a video with $T$ frames $V = [v_1, v_2, \dots, v_T]$, we want to predict per-frame action start probabilities $A = [a_1, a_2, \dots, a_T]$. From the action start probabilities, we then infer the number of action repetitions.   

The proposed network consists of three modules, as shown in Fig.~\ref{fig:racnet}. The \textbf{Feature Encoder} produces per-frame action embeddings and the \textbf{Temporal Repetition Constrain Module (TRC)} forces the learned embeddings to have a self-similarity structure that respects action repetitions in the input video. The \textbf{Action Start Predictor (ASP)} predicts for each frame the probability that an action starts.
In the following, we explain each module in detail.

\subsection{Feature Encoder}

Our encoder is composed of two main parts: spatial feature extractor and local feature aggregator.

\noindent
\textbf{Feature extractor.} We use a pre-trained Video Swin Transformer \cite{liu2021swin} to extract per-frame spatial features for the input video. For each frame $v_i$ in the input video, we extract a feature tensor with dimension $7 \times 7 \times 768$. To retain context information and temporary consistency, we extract features from full-resolution videos. %Note that features are extracted offline during training.

\noindent
\textbf{Local feature aggregator.} To enrich the features with temporal context as in \cite{dwibedi2020counting, hu2022transrac}, we feed the full-resolution spatial features of the input video to a 3D convolution layer with size $3\times3\times3$. Then, we apply a 2D spatial pooling to get the feature embeddings $x_i$ for each frame in the input.

\subsection{Temporal Repetition Constrain Module (TRC)}
\label{sec:scm}

Given the per-frame embeddings $[x_1, \dots, x_T]$ from the encoder network, we construct a temporal self-similarity matrix $S$. The matrix $S$ is a $T\times T$ matrix, where the element $s_{ij}$ represents the similarity between frame $i$ and frame $j$ in the input video. To calculate the similarity score $s_{ij}$, we compute the similarity function $f(i,j)$ between frame $i$ and frame $j$, which is the negative of the Hamming distance between the corresponding frame embeddings, followed by row-wise min-max normalization. 

\noindent
\textbf{Reference TSM.}
%We found that the TSM can be constrained by the prior that makes the feature learning into temporal repetition consistency learning, construct the TSM into a complete repetition structure, so we proposed a method to construct it and name it as reference TSM. 
To generate the reference TSM $S_{ref}$, we assume that frame embeddings of one action must align with frame embeddings of all other repetitions. 
\textit{I.e.}, the embedding of the first frame in one action should have a high similarity with the embedding of the first frame in all repetitions. This also holds for the last frame in each repetition. 
For frames between the start and end of an action, we allow many-to-many alignment as these repetitions might have different durations. To get the final reference TSM matrix, we follow the following steps:
\vspace{2mm}
\begin{enumerate}[itemsep=-3pt]%[nosep]
    \item Define a $T \times T$ matrix and initialize it with zeros, except for the diagonal which is initialized to one.

    \item Set $s_{ij}$ to one if both $i$ and $j$ correspond to either the start of an action repetition or its end.

    \item For each pair of action repetitions, we set all the elements that lie on the line connecting locations $ii'$ and $jj'$ to one, where $i$ is the start of the first repetition and $j$ is its end, where $i', j'$ corresponds to the start and end of the second repetitions.

    \item Smooth the similarity matrix.
\end{enumerate}
\vspace{2mm}

% \YL {REVIEW: Figure 3 visualizes TSM, but lacks a specific and clear explanation of the visualization results.}

\begin{figure}[t]
  \centering
    \begin{subfigure}{0.15\textwidth} %width=2.1cm,height=2.1cm
        \centering
        % \hspace{-1mm}
        \includegraphics[width=3cm]{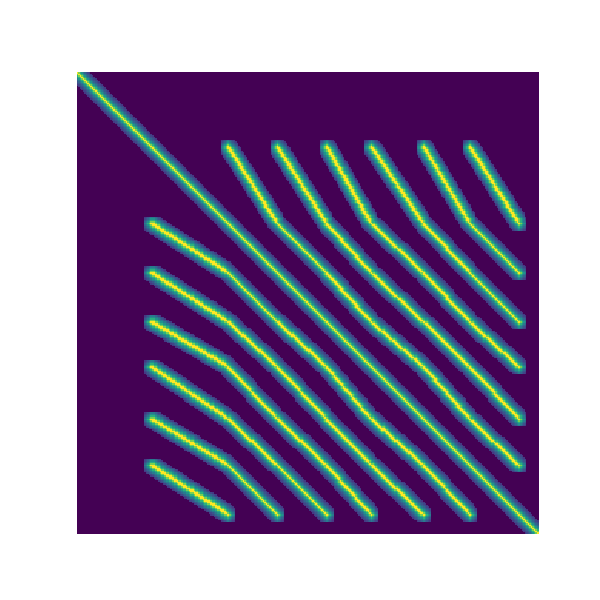}\vspace{-10pt}
        %\caption{} % 
    \end{subfigure}
    \begin{subfigure}{0.15\textwidth}
            \centering
        % \hspace{-1mm}
        \includegraphics[width=3cm]{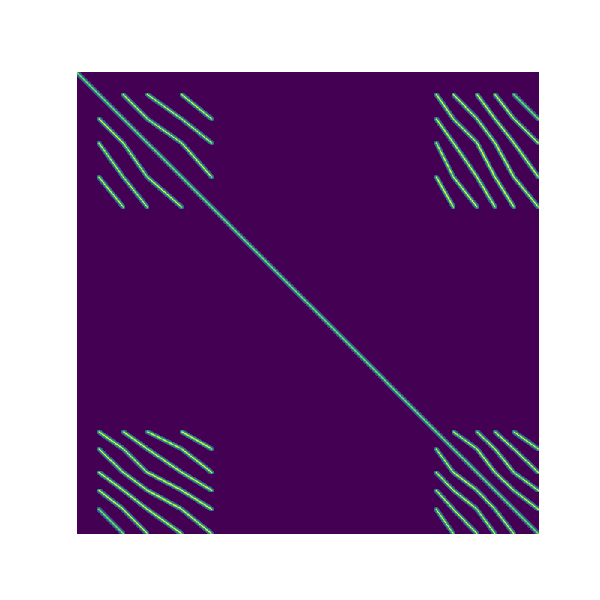}\vspace{-10pt}
        %\caption{Long interruption} % 
    \end{subfigure}
    \begin{subfigure}{0.15\textwidth}
            \centering
        % \hspace{-1mm}
        \includegraphics[width=3cm]{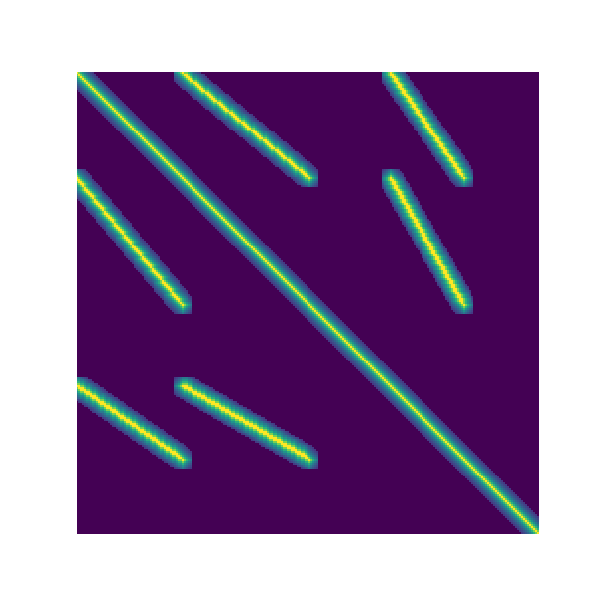}\vspace{-10pt}
        %\caption{Different durations} % 
    \end{subfigure}
  \vspace{-2mm}
  \caption{Examples of generated reference temporal self-similarity matrices for three different videos. Left: The actions start later in the video. Middle: There is a long break until the actions are continued. Right: The actions have different durations.}
  \label{fig:gen-gt-tsm}
  \vspace{-5mm}
\end{figure}

Fig.~\ref{fig:gen-gt-tsm} shows three examples of the produced reference TSM. The matrix depicts the temporal structure of the input video and shows where the repetitions start and end. It also exposes the periods where no actions are carried out in the video. For example, Fig.~\ref{fig:gen-gt-tsm} (a) shows an action with late starts, (b) reveals that there is a long interruption in the action, and (c) illustrates the repetitions with different durations. 
By using such a reference TSM matrix, we soft-constrain the network to learn features that share the self-similarity structure of repeated actions.  
%generated TSM to follow a structure that depicts the number of repetitions in the input video and to retain the temporal repetition consistency among related frames. 

\noindent
\textbf{Temporal repetition consistency loss.} %In contrast to previous approaches \cite{dwibedi2020counting,hu2022transrac,li2023full} that use TSM as the bottleneck in their networks, 
In order to guide the network to learn features where the self-similarity of the features reflects the repetitions of the action, we propose a \textbf{t}emporal \textbf{Re}petition \textbf{Co}nsistency loss (tReCo loss) on the generated TSM $S_{ref}$. Specifically, the sum-of-squared error between the generated TSM $S$ and the reference TSM $S_{ref}$ is defined as an objective function:
%\vspace{-6mm}
\begin{equation}
    %\vspace{-2mm}
    \ell_{\text {tReCo}}(S, S_{ref})= \sum_{i,j=1,1}^{T, T}(S_{ij}-S_{ref_{ij}})^2  \cdot \mathbf{1}_{S_{ref_{ij}}\neq0},
    \label{eq:tsm-loss}
\end{equation}
where we only consider the non-zero elements in $S_{ref}$ since we only want to enforce the self-similarity of features when actions are repeated. For other parts of the video, self-similarities can occur as well, e.g., when a person makes a break and stands around. We thus do not penalize self-similarities where the action is not performed.            

%The reference TSM $S_{ref}$ using the ground-truth annotations of the repetitions start and end frames. Then, we use sum-of-squared error between the generated TSM $S$ and the reference TSM $S_{ref}$ as an objective function

\subsection{Action Start Predictor (ASP)}
The Action Start Predictor (ASP) module takes the frame-wise embeddings $[x_1, \dots, x_T]$ from the encoder network as input, and predicts the probability of each frame being a start frame for an action repetition. 
%This means that our model is not constrained by the TSM bottleneck and has direct access to all frames in the input video.

\noindent
\textbf{Sigmoid-Multi-Stage TCN (SMS-TCN).}
We adopt a variant of the multi-stage temporal convolutional network (MS-TCN) \cite{farha2019ms} for the action prediction. Specifically, the activation function of each stage is replaced by a sigmoid function to output the frame-wise action start probabilities  $[a_1, \dots, a_T]$. As the stage progresses, the predicted probabilities are refined due to increased temporal receptive fields, which enable the assimilation of contextual information across numerous frames. In contrast to MS-TCN, only the output of the last stage is used for the loss calculation. 

% Each stage consists of several layers of dilated temporal convolutions with skip connections, such that the dilation factor is doubled at each layer (see Fig~\ref{fig:dilation-layer}). Using exponentially increasing dilation factors helps the network have a large receptive field and capture the temporal structure in the input video.

\noindent
\textbf{Start prediction loss.} As a target for this module, we define a Gaussian around each frame that corresponds to an action start. 
%Unlike recent research using the Gaussian distribution for each action cycle while existence actions ambiguous in the boundaries, especially when the repetitions are continuous. 
The loss function for the action start prediction is the sum-of-squared error
%\vspace{-2mm}
\begin{equation}
%\vspace{-2mm}
    \ell_{\text {sse}}(A , \hat{A})=\sum_{i=1}^{T}(a_{i}-\hat{a}_{i})^2 ,
    \label{eq:density-loss}
\end{equation}
where $\hat{a}_i$ is the predicted probability of frame $i$ being an action start, and $a$ is the target value.

\noindent

\subsection{Loss}
The final loss function to train our model has two parts
\begin{equation}
    \ell = \ell_{sse} + \lambda \ell_{tReCo},
    \label{total_loss}
\end{equation}
where $\ell_{sse}$ defined in (\ref{eq:density-loss}) is the loss for predicting the action start probabilities, and $\ell_{tReCo}$ defined in (\ref{eq:tsm-loss}) is the temporal self-similarity loss. We set $\lambda = 1.0e^{-5}$ to balance the impact of the two losses.

\subsection{Inference}
During inference, we feed the videos into the encoder network to get the per-frame embeddings, which are fed into the action start predictor (ASP) to get the per-frame predicted probabilities of action start. Note that the Temporal Repetition Constraint Module (TRC) is not used during inference and is only utilized during training. 

To get the number of repetitions of the action in the input video, we need to count the number of frames that correspond to an action start in the predicted output. To get these frames, we first find all the local peaks in the predicted probabilities. Then we only keep those frames with a prominence~\cite{helman2005finest} value higher than a threshold.
%\YL{(measures how much the peak stands out due to its height and its location relative to other peaks)}

%Prominence\cite{helman2005finest} is used in topography to measure the height of mountains. The prominence of a peak quantifies its conspicuousness, which results from a combination of its inherent elevation and its positioning in relation to neighboring peaks. \ie, it measures how much the peak stands out due to its height and its location relative to other peaks. By defining the threshold of prominence, any peaks that have a relatively low prominence will be ignored to eliminate noisy peaks in density maps. Fig.\ref{fig:prominence} shows an example of how prominence is calculated. The prominence of peak 5 is defined as the difference between the height of peak 5 and its neighboring valley e with a higher height (valley e is higher than valley d). By defining the threshold of prominence, any peaks that have a relatively low prominence will be ignored to eliminate noisy peaks in density maps. 

\begin{comment}
\begin{figure}[tb!]
    \centering
    \includegraphics[width=8cm]{prominence_peaks.pdf} 
    \caption[caption]{Prominence of peaks in 1D signal. Take point 6 for example, the prominence is the max(min(b,d), min(h, i)), which reveals how much peak 6 stands out due to its height and its location relative to other peaks.}
    \label{fig:prominence}
\end{figure}
\end{comment}

%% file: sec/4_experiments.tex
\section{Experiments}
\label{sec:exps}

% \YL {REVIEW: a thorough analysis and resolution of the potential changes in computational complexity and the impact on training and inference efficiency. I recommend the authors to provide an analysis of the computational complexity and training/inference efficiency.}
%\YL {REVIEW: The authors claim that they take all the video frames as input. Howerver, the authors did not provide a sufficiently detailed and accurate description in the paper of how an untrimmed long video is inputted into the model. For example, the mean video duration of RepcountA dataset is 30.67s, so that the number of input video frame exceeds one thousand, which is too large to model.}

\subsection{Dataset}

We use three large-scale datasets to evaluate our approach: RepCountA\cite{hu2022transrac}, UCFRep\cite{Zhang_2020_CVPR}, and Countix\cite{dwibedi2020counting}.

\textbf{RepCountA}. It is the largest repetitive action counting dataset with two parts. Part A consists of 1041 videos from YouTube, and part B contains 410 videos recorded in a local school. 
The average duration of videos in part A is 30.67 seconds, which is 4-5 times larger than the other datasets. RepCount is the only dataset that has fine-grained annotations of start and end for each action cycle. Part B is not released yet, so all experiments will be conducted on Part A. For comparisons, we use the test subset which has 152 videos.

\textbf{UCFRep}. %This dataset proposed a new perspective and definition of the cycle to count, and the annotations are count values of videos.
This dataset provides the number of actions for each video. 
All the videos are taken from UCF101~\cite{soomro2012ucf101}. The average duration of the videos is 8.15 seconds. For comparisons, we use the validation set which has 105 videos. 

\textbf{Countix}. This dataset is a subset of Kinetics~\cite{kay2017kinetics} and only contains the segments of repeated actions with the corresponding count annotations. The test data originally consisted of 2719 videos, but only 1692 videos are still available. We report all the results on the 1692 videos. The average duration of the videos is 6.13 seconds. 

\subsection{Evaluation Metrics}

For evaluation, we use the common metrics used by state-of-the-art approaches~\cite{dwibedi2020counting, hu2022transrac}: Off-By One Accuracy (OBOA) and Mean Absolute Error (MAE).

\noindent
\textbf{Off-By-One Accuracy (OBOA)}. If the difference between the predicted count and the ground truth is less than or equal to 1, the prediction is considered as correct, otherwise as wrong. 
%The OBO accuracy is the classification tolerance by one accuracy.

\noindent
\textbf{Mean Absolute Error (MAE)}. This metric calculates the absolute difference between the predicted count and the ground-truth count, normalized by the ground-truth count.

\subsection{Implementation Details}

%We implement our approach using Pytorch\cite{paszke2017automatic}. 
For feature extraction, we use Video Swin Transformer~\cite{liu2021swin} pre-trained on Kinetics400~\cite{kay2017kinetics}. As in previous works, it is frozen for a fair comparison. We train our model with a learning rate of $6.4 \times 10^-5$ and use ADAM optimizer with a batch size of 16. We trained our models for 100 epochs on an Nvidia RTX A6000 with 48GB memory in less than 9 hours. Inference takes 6 seconds per video on average. The prominence\footnote{Please refer to supplementary material for more details: \href{https://sigport.org/sites/default/files/docs/ICIP24\_RACnet\_supp.pdf}{https://sigport.org/sites/default/files/docs/ICIP24\_RACnet\_supp.pdf}} threshold is set to 0.2.
%The prominence threshold is 0.3 for training and 0.2 for testing.

\subsection{Comparision with the State-of-the-Art}

We compare the proposed approach with state-of-the-art approaches on the RepCountA~\cite{hu2022transrac} dataset in Table~\ref{tab:performace-repcountA}. Transrac~\cite{hu2022transrac}\footnote{\href{https://github.com/SvipRepetitionCounting/TransRAC}{https://github.com/SvipRepetitionCounting/TransRAC}} and ME-Rac~\cite{qiu2024multipath}\footnote{\href{https://github.com/yicheng-2019/ME-RAC}{https://github.com/yicheng-2019/ME-RAC}} do not compute MAE and OBOA based on the count of action repetitions, but on the density map that is generated from the ground-truth. Since a comparison based on density maps is not accurate and does not allow to compare methods that use a different temporal resolution, we report the result of Transrac using the standard protocol proposed by~\cite{dwibedi2020counting}. The results show a major difference in MAE, but a small difference in OBOA if the protocol is changed. The approaches~\cite{li2023full, qiu2024multipath} are based on TransRac. When comparing to methods using the same protocol, our approach outperforms the state of the art by a large margin. The concurrent work \cite{qiu2024multipath} achieves a lower MAE and OBOA, but the numbers are not comparable due to the different protocols. We also include the average inference time for a video for methods with available source code. Our method is slower than \cite{dwibedi2020counting, hu2022transrac} since it uses the full temporal resolution. We expect that the concurrent work~\cite{qiu2024multipath} is much slower than our approach since it uses two stages model, where objects are first detected in each video frame.

% While our approach achieves comparable MAE to other methods, it outperforms state-of-the-art approaches with respect to OBOA by a large margin of roughly 6.5\%. As OBOA is a more strict metric, this highlights that our approach is more effective in capturing the repetitive action nature and predicting the correct number of repetitions.

\begin{table}[t]
    \centering
    \resizebox{0.8\linewidth}{!}{
    \begin{tabular}{llll}
        \toprule
        Method & MAE$\downarrow$ & OBOA$\uparrow$ & Infer Time (s) \\
        \midrule
        TransRac*~\cite{hu2022transrac} & 0.4431 & 0.2913 & 1.1194 \\       
        Li et al.*~\cite{li2023full} & 0.4103 & 0.3267 & -\\
        ME-Rac*~\cite{qiu2024multipath} & \textit{\textbf{0.3529}} & \textit{\textbf{0.4018}} & -\\
        \midrule
        RepNet~\cite{dwibedi2020counting} & 0.9950 & 0.0134 &  0.4656\\
        Zhang et al.~\cite{zhang2020context} & 0.8786 & 0.1554 & - \\
        TransRac~\cite{hu2022transrac} & 0.6099 & 0.2763 & 1.1194 \\
        % ours           & 0.5661 & \textbf{0.3333}\\
        RACnet (ours) & \textbf{0.4441} & \textbf{0.3933} & 6.1689 \\
        \bottomrule
    \end{tabular}
    }
    \vspace{-2mm}
    \caption{Comparison to state-of-the-art approaches on the RepCountA dataset. * denotes a different evaluation protocol.        
    % Note that the labeled count of Transrac\cite{hu2022transrac}\tablefootnote{https://github.com/SvipRepetitionCounting/TransRAC} is computed as the sum of ground-truth density map, which is biased with other works. For fair comparison, we report the result of Transrac with the same protocol as in RepNet (marked with *). Li et al.~\cite{li2023full} is based on TransRac and we assume that the results also have the same issue.
    } 
    \label{tab:performace-repcountA}
\end{table}

We also compare the generalization performance of our approach with other methods. In this setup, all the models are trained on the training set of RepCountA~\cite{hu2022transrac} and evaluated on UCFRep~\cite{Zhang_2020_CVPR} and Countix~\cite{dwibedi2020counting}. As shown in Table~\ref{tab:performace-dataset}, our model generalizes very well to unseen videos on both datasets.

\begin{table}[t]
    \centering
    \resizebox{0.85\linewidth}{!}{%
    \begin{tabular}{lllll}
        \toprule
                    & \multicolumn{2}{l}{UCFRep} & \multicolumn{2}{l}{Countix} \\
        \midrule
        Method        & MAE$\downarrow$        & OBOA$\uparrow$      & MAE$\downarrow$        & OBOA$\uparrow$       \\
        \midrule
        Li et al.*~\cite{li2023full} &  \textit{\textbf{0.4608}}	& 0.3333 & - & - \\
        \midrule
        RepNet~\cite{dwibedi2020counting}  & 0.9985     & 0.0090   &   0.8441   &  0.1600 \\ 
        % Zhang et al.~\cite{zhang2020context} & 0.762 & \textbf{0.412} & - & - \\
        TransRac~\cite{hu2022transrac}  & 0.6401   & 0.3240  &  0.5804 & 0.3782   \\
        % ours           & 0.4975 & 0.3429 & \textbf{0.5278} & \textbf{0.3859}  \\
        RACnet (ours)       &  \textbf{0.5260} & \textbf{0.3714} &  \textbf{0.5278} & \textbf{0.3924} \\

        \bottomrule
    \end{tabular}
    }
    \vspace{-2mm}
    \caption{Generalization on the UCFRep and Countix datasets. All models are trained on the training set of RepCountA. * denotes a different evaluation protocol.}
    \label{tab:performace-dataset}
\end{table}

\begin{table}[b]
     \centering
     \resizebox{0.65\linewidth}{!}{
     \begin{tabular}{llll}
         \toprule
                  & MAE$\downarrow$        & OBOA$\uparrow$             \\
         \midrule
         w/o tReCo loss    &  0.4571 & 0.3667   \\ 
         w tReCo loss  & \textbf{0.4441} & \textbf{0.3933} \\
        \bottomrule
     \end{tabular}
     }
     \vspace{-2mm}
     \caption{Impact of the tReCo loss on the RepCountA dataset.}
     \label{tab:ablation-tsmLoss}
 \end{table}

\subsection{Ablation Studies}
\noindent
\textbf{Impact of the tReCo loss.} 
%Our model is trained using two losses: tReCo loss and squared error for predicting the action start probabilities. In this section, we analyze the impact of the tReCo loss. 
Table~\ref{tab:ablation-tsmLoss} shows the results of our model trained with and without the tReCo loss on the RepCountA dataset. Using the tReCo loss improves the results with respect to both metrics. This shows the benefit of adding an additional loss to ensure that the self-similarity of the features is consistent with the structure of the repeated actions. 

\noindent
\textbf{Impact of the similarity measure.}
To calculate the predicted TSM, we use the negative Hamming distance between pairs of frame embeddings. In Table~\ref{tab:ablation-tsm-calculate}, we compare it to other variants such as Euclidean distance, correlation, and self-attention. The Hamming distance performs best.

\begin{table}[t]
    \centering
    \resizebox{0.65\linewidth}{!}{%
    \begin{tabular}{llll}
        \toprule
        Distance       & MAE$\downarrow$        & OBOA$\uparrow$      \\
        \midrule
        Euclidean distance   & 0.5661 & 0.3333  \\ 
        Correlation & 0.6307  & 0.2200 \\
        Self-attention & 0.7459  & 0.1730  \\
        Hamming distance & \textbf{0.4441} & \textbf{0.3933} \\
        \bottomrule
    \end{tabular}
    }
    \vspace{-2mm}
    \caption{Impact of different similarity/distance measures for calculating the TSM on the RepCountA dataset.}
    \label{tab:ablation-tsm-calculate}
\end{table}

\begin{table}[t]
     \centering
     \resizebox{0.45\linewidth}{!}{
     \begin{tabular}{llll}
         \toprule
         Stride         & MAE$\downarrow$        & OBOA$\uparrow$           \\
         \midrule
          %4  & 0.5254  & 0.3333   \\ 
          3 & 0.5216 & 0.3467  \\
          2 & 0.5420 & 0.3133 \\
          1 & \textbf{0.4441} & \textbf{0.3933} \\
         \bottomrule
     \end{tabular}
     }
     \vspace{-2mm}
     \caption{Impact of the temporal resolution on the RepCountA dataset.}
     \label{tab:ablation-stride}
 \end{table}

\noindent
\textbf{Impact of temporal resolution.}
In Table~\ref{tab:ablation-stride}, we evaluate the impact of the temporal resolution. By increasing the sampling stride, we decrease the temporal resolution. The results show that using the full temporal resolution, i.e., stride 1, gives a substantial improvement for all metrics. 
 
\begin{table}[b]
\centering
\resizebox{0.8\linewidth}{!}{%
    \centering
    \begin{tabular}{lllll}
        \toprule
        Dataset       & Methods & MAE$\downarrow$        & OBOA$\uparrow$     \\
        \midrule
        \multirow{4}{*}{RepCountA} & Li et al.*~\cite{li2023full} & \textit{\textbf{0.4366}} & 0.3000  \\
        & RepNet~\cite{dwibedi2020counting} & 0.8283 & 0.2933 \\
                                   & Transrac~\cite{hu2022transrac} & 0.5064 & 0.1866 \\
                                   & RACnet (ours) & \textbf{0.4441} & \textbf{0.3933}  \\            
        \midrule              
        \multirow{3}{*}{Countix} &  RepNet~\cite{dwibedi2020counting} & 0.6028 & 0.1158  \\
                                  & Transrac~\cite{hu2022transrac} & 0.5483 & 0.3712  \\
                                  & RACnet (ours) & \textbf{0.5278} & \textbf{0.3924} \\

        \midrule
        \multirow{3}{*}{UCFRep} & RepNet~\cite{dwibedi2020counting}   & 0.6654 & 0.2476  \\
                                & Transrac~\cite{hu2022transrac} & 0.5987 & 0.2952  \\
                                & RACnet (ours)   & \textbf{0.5260} & \textbf{0.3714}  \\
        
        \bottomrule
    \end{tabular}
    }
    \vspace{-2mm}
    \caption{Impact of using full temporal resolution of the videos on the RepCountA, Countix and UCFRep dataset.}
    \label{tab:ablation-full-resolution}
\end{table}

\noindent
\textbf{Impact of full resolution.}
%To our knowledge, we are the first to use full resolution for the repetitive action counting task, both during training and testing. 
Previous methods either sub-sample the input frames or fix the number of input frames to a few frames. We compare the performance of our approach with other methods when the full resolution of the videos is used in Table \ref{tab:ablation-full-resolution}. Compared to Tables~\ref{tab:performace-repcountA} and~\ref{tab:performace-dataset}, using the full resolution improves the performance of RepNet but deteriorates the performance of TransRac and \cite{li2023full}. Even if the full temporal resolution is used by all methods, our approach outperforms these approaches on all datasets.

\noindent\textbf{Predicting action start vs.\ periodicity.}
Our approach predicts the start of an action. In Table~\ref{tab:ablation-pvs-preda}, we compare it to predicting the periodicity with our approach.
While the start of an action is defined by a Gaussian with very small variance, the periodicity is less peaked.   
%Compared to our peak action start prediction, periodicity based method apply a smooth Gaussian distribution to every action, leading to worse performance. 
The results show
that predicting the start of the action is better than the periodicity.
% We compare predicting action start with periodicity adopted in the previous methods. For the comparison we supersede the prediction of action start by the periodicity, train our model on RepCountA\cite{dwibedi2020counting}, and test on the same dataset. Table ~\ref{tab:ablation-pvs-preda} shows that predicting the start of the action is better than the periodicity when applying a full-resolution model.  

 \begin{table}[t]
     \centering
     \resizebox{0.6\linewidth}{!}{
     \begin{tabular}{llll}
         \toprule
         Prediction  & MAE↓        & OBOA↑           \\
         \midrule
           periodicity  &  1.0510 & 0.1667   \\ 
           action start & \textbf{0.4441} & \textbf{0.3933} \\
         \bottomrule
     \end{tabular}
     }
     \vspace{-2mm}
     \caption{Comparison of predicting action start and periodicity on the RepCountA dataset.}
     \label{tab:ablation-pvs-preda}
 \end{table}
 
%\vspace{-10mm}

\subsection{Visualization} 

% To inspect the learned embeddings by our network, we reduce the dimensionality of the per-frame embeddings using principal component analysis (PCA) and visualize the first component. This process exposes captivating quasi-sinusoidal patterns exhibited by the embeddings over time. By visually inspecting frames at the points where the embeddings change direction, we observe that the retrieved frames portray the person or object in similar states but at different time intervals.

%\vspace{-1mm}
\begin{figure}[b]
  \centering
  \vspace{-3mm}
    \begin{subfigure}{0.1\textwidth}
        \includegraphics[width=2cm]{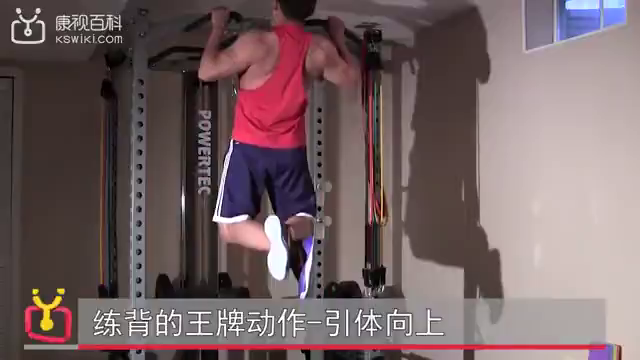}\vspace{-10pt}
        \\
        \\
        % \includegraphics[width=2.5cm]{figures/vis-tsm-png/stu2_55_44.png}
        % \\
        % \\
        % \\
        % \\
        % \includegraphics[width=2.5cm]{figures/vis-tsm-png/stu4_29_143.png}
        % \\
        \vspace{-5mm}
        \quad\caption{}
      \end{subfigure}
    \begin{subfigure}{0.1\textwidth}
        \includegraphics[width=2cm,height=2cm]{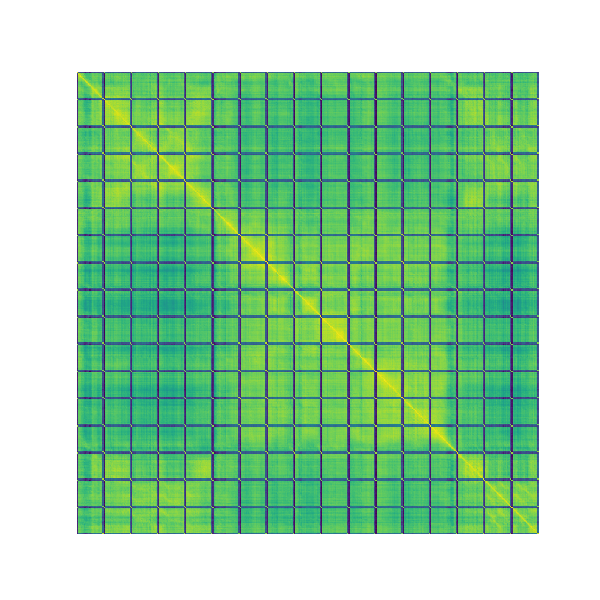}\vspace{-10pt}
        \\
        % \includegraphics[width=2.5cm,height=2.5cm]{figures/vis-tsm-png/stu2_55tsm.png}
        % \\
        % \includegraphics[width=2.5cm,height=2.5cm]{figures/vis-tsm-png/stu4_29tsm.png}
        \vspace{-5mm}
        \quad\caption{}
      \end{subfigure}
    \begin{subfigure}{0.1\textwidth}
        \includegraphics[width=2cm,height=2cm]{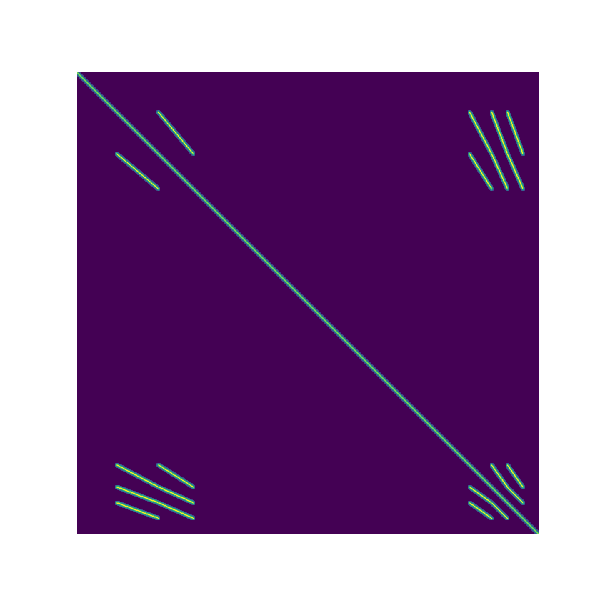}\vspace{-10pt}
        \\
        % \includegraphics[width=2.5cm,height=2.5cm]{figures/vis-tsm-png/stu2_55_gt_TSM.png}
        % \\
        % \includegraphics[width=2.5cm,height=2.5cm]{figures/vis-tsm-png/stu4_29_gt_TSM.png}
        \vspace{-5mm}
        \caption{}
      \end{subfigure}
    \begin{subfigure}{0.1\textwidth}
    \includegraphics[width=2cm,height=2cm]{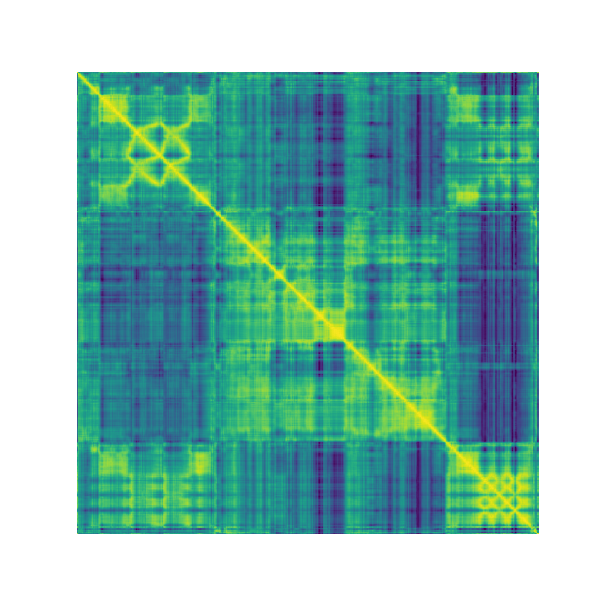}\vspace{-10pt}
    \\
    % \includegraphics[width=2.5cm,height=2.5cm]{figures/vis-tsm-png/stu2_55_pred_tsm.png}
    % \\
    % \includegraphics[width=2.5cm,height=2.5cm]{figures/vis-tsm-png/stu4_29_pred_tsm.png}
    \vspace{-5mm}
    \quad\caption{}
  \end{subfigure}

  \caption{Visualizations of TSMs. (a) Example video. (b) TSM of RepNet. (c) Reference TSM. (d) TSM of our approach. Yellow indicates high similarity and blue stands for low similarity.  %The TSM loss improve the performance of our model and leads the TSM into a more distinct similarity representation.}
  }
  \label{fig:vis_tsm}
\end{figure}

\noindent
\textbf{Visualizations of TSM.}
% reference TSM, repnet, without tsm, with tsm
Fig.~\ref{fig:vis_tsm} shows visualizations of the temporal self-similarity matrix (TSM). Fig.~\ref{fig:vis_tsm} (a) is a video from RepCountA with a long break between the actions. Different from the TSM of RepNet in (b), the TSM of our approach (d) is trained to capture the repetitive action structure (c).

\begin{figure}[t]
    \centering
    \includegraphics[width = 0.45\textwidth]{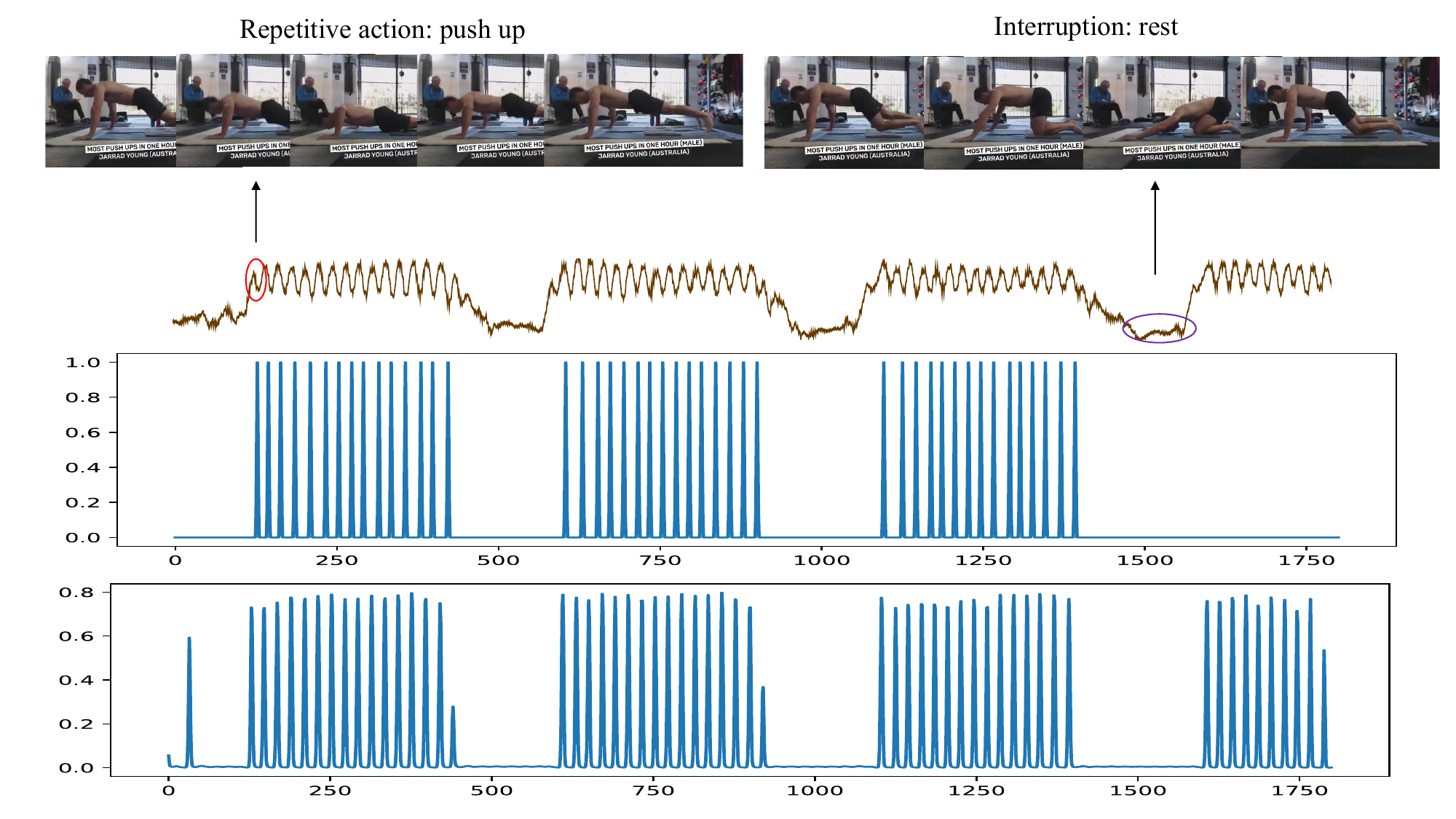} 
    \caption[caption]{Example of incorrect annotation. From top to bottom: several key frames, 1D PCA of feature embeddings, ground truth action start, predicted action start probabilities.}
    \label{fig:vis_video}
\end{figure}

\noindent
\textbf{Visualizations of action start probabilities.}
% groudtruth density, transrac, racnet
Fig.~\ref{fig:vis_video} shows an example from RepCountA~\cite{hu2022transrac}. The video contains 4 sessions of push-ups. The first three sessions contain 15 push-ups and the last session, which was not annotated, contains 10 push-ups. Except for a wrong peak at the beginning, our approach recognizes each push-up.     
We show the 1D PCA of per-frame feature embeddings, the action start annotations of the ground truth, and the predicted action start probabilities of our approach, which results in 55 counted repetitions. The 1D PCA shows the regular changes in actions and finds the breaks between the repetitions. For comparison, RepNet predicts 17 and TransRac predicts 14 repetitions for this video.

\begin{figure}[t]
    \centering
    \includegraphics[width = 0.5\textwidth]{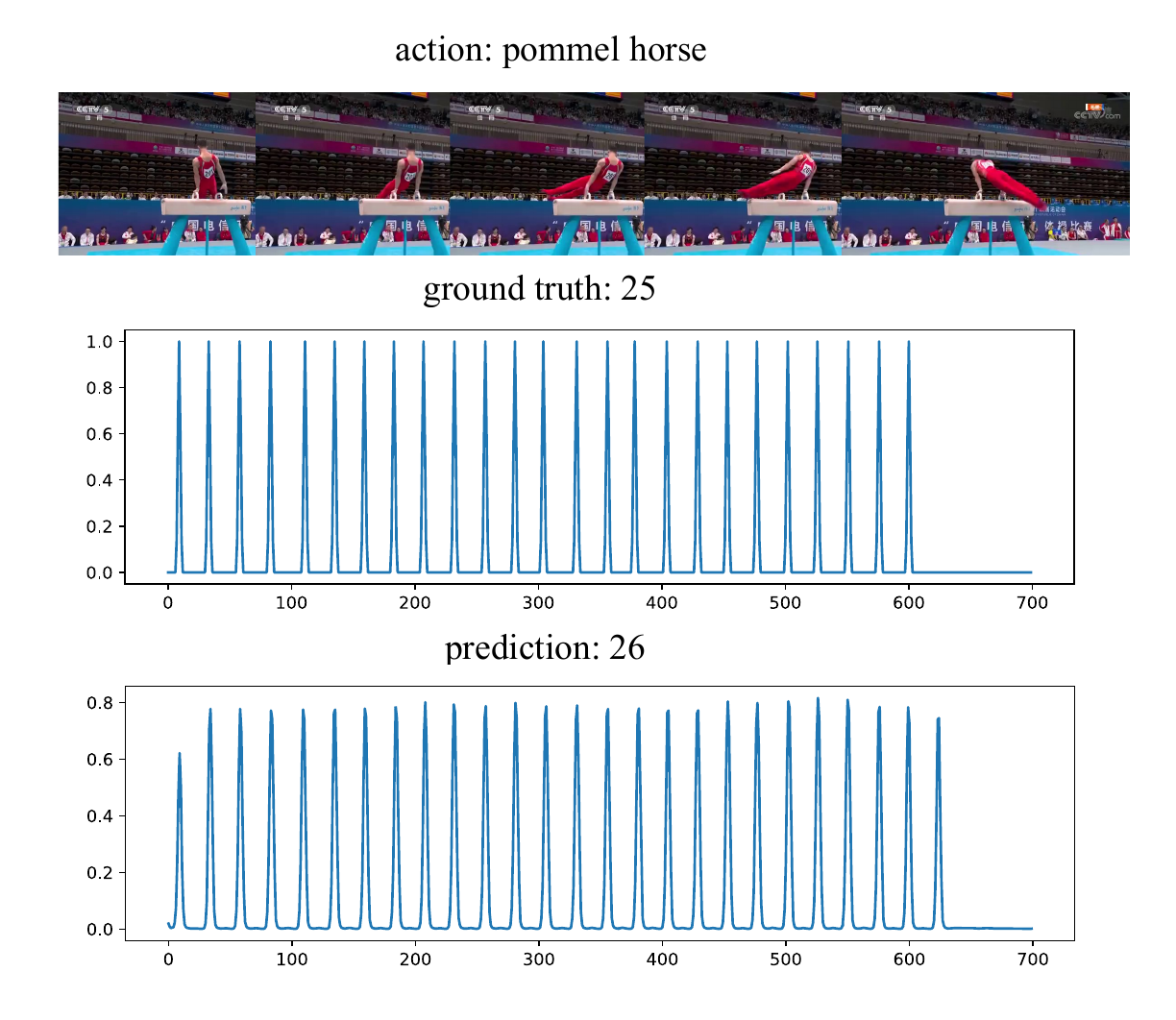} 
    \vspace{-8mm}
    \caption[caption]{Example of predicted action start probabilities. Note that the last action is missing in the annotation but visible in the pommel horse video. }
    \label{fig:vis_correct1}
\end{figure}

\noindent
\textbf{Example of prediction.}
Fig.~\ref{fig:vis_correct1} shows an example of action start predictions for a video from RepCountA. Our approach successfully localizes the start of each pommel horse action and predicts the correct number of action repetitions. Note that the last action is missing in the annotation but visible in the video.

% \YL {online website to show the videos?}
% \noindent
% We also provide the qualitative results with the example videos which show the original video and at the bottom the predicted action start probabilities. 

%% file: sec/5_conclusion.tex
%\vfill
%\pagebreak

\begin{comment}
\section{RELATION TO PRIOR WORK}
\label{sec:prior}

The text of the paper should contain discussions on how the paper's
contributions are related to prior work in the field. It is important
to put new work in  context, to give credit to foundational work, and
to provide details associated with the previous work that have appeared
in the literature. This discussion may be a separate, numbered section
or it may appear elsewhere in the body of the manuscript, but it must
be present.

You should differentiate what is new and how your work expands on
or takes a different path from the prior studies. An example might
read something to the effect: "The work presented here has focused
on the formulation of the ABC algorithm, which takes advantage of
non-uniform time-frequency domain analysis of data. The work by
Smith and Cohen \cite{Lamp86} considers only fixed time-domain analysis and
the work by Jones et al \cite{C2} takes a different approach based on
fixed frequency partitioning. While the present study is related
to recent approaches in time-frequency analysis [3-5], it capitalizes
on a new feature space, which was not considered in these earlier
studies."
\end{comment}

\section{Conclusion}
\label{sec:conc}
In this paper, we proposed a framework 
%for temporal repetition consistency learning
for repetitive action counting in both short and long videos. Our framework casts the problem into action start prediction and calculates the number of actions by counting the number of frames that correspond to a repetition start. In contrast to previous approaches, we feed the full-resolution sequences to our model and do not use the temporal similarity matrix (TSM) as an intermediate representation. Instead, we proposed a temporal repetition constrain loss that forces the learned frame-wise embeddings to capture the repetitive consistency of the action. The proposed loss improves the accuracy of predicting action counts. The proposed framework achieves state-of-the-art results on three datasets.

\bibliographystyle{IEEEbib}
\bibliography{strings,refs}